\documentclass{article} % For LaTeX2e

\usepackage{iclr2027_conference,times}

\usepackage{amsmath,amsfonts,bm}

\def\eqref#1{equation~\ref{#1}}
\def\1{\bm{1}}

\DeclareMathAlphabet{\mathsfit}{\encodingdefault}{\sfdefault}{m}{sl}
\SetMathAlphabet{\mathsfit}{bold}{\encodingdefault}{\sfdefault}{bx}{n}

\usepackage{hyperref}
\usepackage{url}
\usepackage{xcolor}
\usepackage[normalem]{ulem}
\usepackage{booktabs}
\usepackage{amsmath}
\usepackage{amssymb}
\usepackage{mathtools}
\usepackage{amsthm}
\usepackage{enumitem}
\usepackage{multirow}

\usepackage[table]{xcolor}
\definecolor{basepurple}{HTML}{bdb2ff}
\definecolor{CMTgreen}{RGB}{0,120,0}
\definecolor{lightgreen}{RGB}{220,255,220}
\definecolor{lightred}{RGB}{255,220,220}
\usepackage{graphicx}
\usepackage{threeparttable}

\theoremstyle{plain}

\theoremstyle{definition}

\theoremstyle{remark}

\newcommand{\NOTE}[1]{}   % editorial notes: rendered as nothing

\newcommand{\Succ}{\textsc{Succ}}
\title{Trajectory Learnability for Offline On-Policy Distillation with Imperfect Teachers}

\author{%
\makebox[\dimexpr\textwidth-2\tabcolsep\relax][c]{%
\begin{tabular}[t]{c}
\rule{0pt}{32pt}% 控制标题与作者之间的距离
\textbf{Yihao Ai}$^{1,2}$%
\thanks{Work done during an internship at ByteDance.}
\hspace{1.2em}
\textbf{Beibei Lin}$^{2}$
\hspace{1.2em}
\textbf{Weilong Yan}$^{1}$
\hspace{1.2em}
\textbf{Huaijian Zhang}$^{2}$
\\[0.35em]
{\normalfont
$^{1}$National University of Singapore
\hspace{2em}
$^{2}$ByteDance}
\\[0.25em]
{\normalfont\small
\texttt{\{yihao,yanweilong\}@u.nus.edu}}
\\[-0.05em]
{\normalfont\small
\texttt{\{beibei.lin,huaijian.zhang\}@bytedance.com}}
\end{tabular}%
}%
}

\iclrfinalcopy % arXiv only: reveal authors and disable review line numbers.
\begin{document}

\maketitle
% arXiv preprint: suppress the ICLR camera-ready header.
\fancyhead{}
\pagestyle{plain}

\begin{abstract}
Offline on-policy distillation gains efficiency by collecting student trajectories and teacher supervision once and reusing them throughout optimization. The same reuse makes imperfect supervision persistent. Since even strong teachers can fail, we ask \emph{what remains learnable from imperfect teacher supervision?}
Teacher failure is only a coarse problem-level signal and does not imply that all supervision along the associated student trajectory is unhelpful.
A natural alternative is to estimate teacher recoverability along the trajectory, but repeated continuations largely erase the efficiency advantage of offline distillation.
We instead use teacher-successful problems to define a cheap reference for what the student can learn. We train on teacher-successful problems and measure how the likelihood of each observed token in trajectories from teacher-failed problems changes.
We use these signed likelihood changes as an operational \emph{learnability signal}: larger increases indicate behavior more strongly promoted by successful-only learning.
We aggregate this signal into trajectory-level weights for the original distillation loss.
Unlike continuation-based estimates, our learnability requires no additional generation and can be computed once from stored trajectories and model checkpoints.
Across mathematical reasoning and code generation, our method improves an offline OPD baseline by up to 2.7 percentage points and matches or outperforms online OPD variants on multiple benchmarks. Despite the additional successful-only distillation stage, it uses 2 GPUs and about 22 GPU hours, compared with 3 GPUs and 36--48 GPU hours for representative online OPD methods.

\end{abstract}

\section{Introduction}

On-policy distillation (OPD) trains a student on states induced by its own generations while using a stronger teacher to provide dense token-level supervision \citep{agarwal2024policy}. Its offline variant improves efficiency by collecting student trajectories and teacher supervision once and reusing them throughout optimization \citep{wu2026lightning}. Yet the same reuse also makes imperfect supervision persistent. Once unreliable teacher guidance enters the offline training set, it can repeatedly shape the student over many optimization steps. Since even strong teachers can fail, this raises a fundamental question for offline OPD: \emph{what remains learnable from imperfect teacher supervision?}

\begin{figure}[t]
\centering
\includegraphics[width=\linewidth]{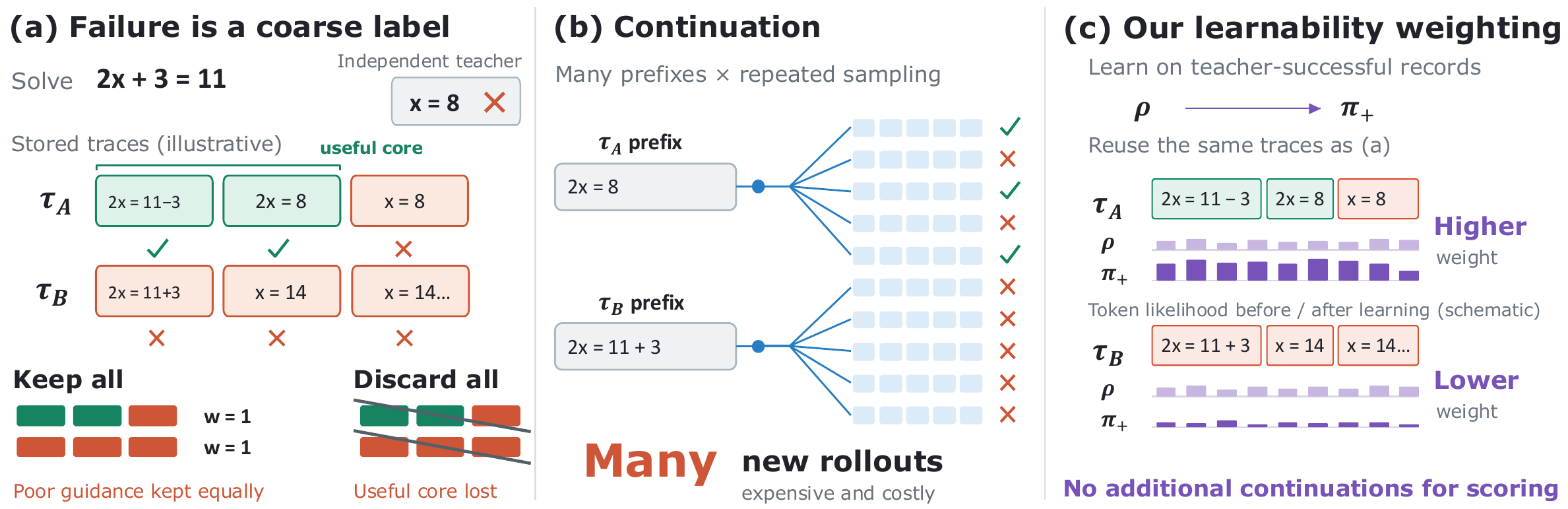}
\caption{Motivation for learnability-weighted OPD.
(a) A failed teacher response provides only a problem-level label. Student trajectories with the same label may deserve different amounts of OPD supervision. Keeping all weights them equally, while discarding all can remove useful supervision.
(b) Estimating recoverability requires repeated continuation and verification from many prefixes.
(c) We compare stored-token likelihoods before and after successful-only learning to assign positive trajectory weights. This requires no repeated continuation rollouts from multiple prefixes.}
\label{fig:teaser}
\end{figure}

A natural first answer is to use whether the teacher succeeds or fails on the underlying problem. Outcome-based filtering uses task success as a coarse signal for selecting reasoning trajectories \citep{zelikman2022star,yuan2023scaling}. In OPD, however, teacher outcome and distillation supervision are distinct: the former is obtained from a generated teacher response, whereas the latter is provided on student-visited states. A failed teacher rollout therefore does not imply that every conditional teacher target on the student trajectory is useless. Discarding teacher-failed cases may remove useful supervision, while retaining them with full weight ignores the information carried by teacher failure. Moreover, student trajectories with the same teacher-failure outcome can exhibit different reasoning behavior. As illustrated in Figure~\ref{fig:teaser}(a), treating records identically cannot capture this variation. Our question is therefore not simply whether teacher-failed supervision should be kept or discarded, but how much should be retained for each trajectory.

Ideally, we would distinguish such trajectories by how much useful guidance remains recoverable along them. One could repeatedly continue from different prefixes and measure whether the teacher can still recover a successful solution. Richer supervision of this form is related to verifier- and process-supervision methods \citep{cobbe2021training,lightman2024let}, while recent OPD methods also model teacher reliability more explicitly \citep{gan2026teacher,zhu2026reorder,zhang2026verify}. In practice, however, recoverability estimation requires repeated generation from many prefixes. At the granularity required to characterize long reasoning trajectories, this cost largely defeats the efficiency advantage of offline distillation, as shown in Figure~\ref{fig:teaser}(b). We therefore seek a cheaper signal of how much useful supervision remains in each teacher-failed trajectory.

Our idea is to train the student on teacher-successful problems and use the update as a reference, which we call \emph{successful-only learning}. We measure how this update changes the likelihood of each observed token in trajectories from teacher-failed problems, and use these signed changes as an operational token-level \emph{learnability signal}. Positive changes indicate behavior promoted by successful-only learning, with larger changes providing stronger evidence of learnability. This signal can weight tokens, be aggregated over spans, or be summarized at the trajectory level. We empirically find trajectory-level weighting most effective, as it adjusts the supervision assigned to a trajectory while preserving the token-level OPD structure. This choice is consistent with selective OPD work showing that trajectory-level weighting can materially affect training \citep{lin2026renio}.

To obtain a stable trajectory-level weight, we first smooth the learnability signal to suppress token-level fluctuations, then normalize the trajectory statistics relative to teacher-failed examples. Rather than relying on differences in raw scores, we use the normalized learnability scores to form coarse low-, mid-, and high-learnability groups and assign each group a bounded positive weight. We refer to the resulting method as \emph{Learnability-Weighted Distillation (LWD)}.

We evaluate our method on mathematical reasoning and code generation, distilling Qwen3-8B into Qwen3 students at two scales. With Qwen3-4B-Base, our method improves the offline OPD baseline from $66.4\%$ to $69.1\%$ on AIME24, from $40.5\%$ to $42.1\%$ on HMMT25, and from $42.7\%$ to $44.5\%$ on LiveCodeBench v5, with gains on the remaining benchmarks and at the 1.7B scale. It also matches or exceeds online OPD variants on multiple evaluations. Despite the additional successful-only distillation stage, the pipeline requires only $2$ GPUs and approximately $22$ GPU hours, compared with $3$ GPUs and $36$--$48$ GPU hours for online OPD methods.

Our contributions are threefold:
\begin{itemize}[leftmargin=1.2em,itemsep=0.15em,topsep=0.25em]

\item We identify the problem of learning from imperfect teacher supervision in offline OPD and introduce a cheap, operational \emph{learnability} signal based on successful-only learning, avoiding the repeated prefix-level continuation sampling required by direct recoverability estimation.

\item We develop a trajectory-level weighting scheme that aggregates local learnability signals into coarse tiers. It reallocates supervision across trajectories while preserving the token-level OPD structure. This makes the weighting more stable and less sensitive to local fluctuations.

\item We demonstrate gains across mathematical reasoning and code generation at two student scales. Our method improves offline OPD and matches or exceeds online selective OPD on multiple benchmarks. It also retains substantially lower computational cost in both GPU hours and usage.

\end{itemize}

% =============================================================================
\section{Related Work}
% =============================================================================

\subsection{Knowledge Distillation and On-Policy Distillation}

Knowledge distillation transfers knowledge from a stronger teacher through predictive distributions, sequence-level supervision, and self-distillation \citep{hinton2015distilling,kim2016sequence,furlanello2018born}. For language models, distillation has further expanded to explanations and chain-of-thought reasoning \citep{hsieh2023distilling,shridhar2023distilling,li2023symbolic,mukherjee2023orca,feng2024keypoint,li2024mode,gu2024minillm}, with broader developments reviewed in \citet{yang2025survey}. On-policy distillation trains the student on states induced by its own generations \citep{agarwal2024policy}, with later work studying its optimization behavior in reasoning models \citep{li2026rethinking}. Lightning-OPD moves this paradigm offline by reusing precomputed student trajectories and teacher supervision \citep{wu2026lightning}. Reusing a fixed collection makes supervision quality particularly consequential, since an unreliable target can influence the student across many optimization steps. We build on this setting and study what remains worth learning when the reused teacher supervision is imperfect.

\subsection{Selective and Reliability-Aware On-Policy Distillation}

Recent OPD methods increasingly treat teacher supervision selectively rather than uniformly. ReNIO reweights negative student trajectories using student--teacher probability information \citep{lin2026renio}, while FiRe-OPD combines trajectory filtering with finer-grained reweighting \citep{li2026filter}. Position-weighted self-distillation studies teacher reliability across reasoning positions \citep{liu2026teacher}, and ExOPD modifies the standard OPD objective to encourage extrapolation beyond direct imitation \citep{yang2026learning}. Outcome-aware methods go further: RA-OPD checks alignment between teacher-induced updates and trajectory reward \citep{gan2026teacher}, ReOrder-OPD uses a proxy for teacher continuation reliability to order prompts \citep{zhu2026reorder}, and TGOPD uses additional teacher probes to gate between OPD and verifier-guided optimization \citep{zhang2026verify}. These approaches differ in their supervision signals and whether they change example selection, training order, or the objective. Offline OPD additionally requires economical signal computation. We ask what remains learnable after teacher failure is observed, using teacher-successful problems as a low-cost reference for reweighting offline OPD supervision.

\subsection{Outcome Supervision and Adaptive Reweighting}

Reasoning post-training often relies on coarse outcome signals through self-training, rejection sampling, and verifier-based selection \citep{zelikman2022star,cobbe2021training,yuan2023scaling}, while process supervision provides denser feedback within reasoning traces \citep{lightman2024let, sang2026crisp, xu2026tip}. Recent work also shows that unsuccessful trajectories can still contain reusable structure that binary rejection discards \citep{deng2026beyond}. More broadly, adaptive example weighting has been studied in curriculum learning, self-paced learning, and robust learning under noisy supervision \citep{DBLP:conf/icml/BengioLCW09,DBLP:conf/nips/KumarPK10,DBLP:conf/icml/JiangZLLF18,DBLP:conf/icml/RenZYU18,DBLP:conf/nips/HanYYNXHTS18}, with related ideas in advantage-weighted and offline policy learning \citep{DBLP:journals/corr/abs-1910-00177,DBLP:conf/nips/KumarZTL20,DBLP:conf/iclr/KostrikovNL22}. Teacher outcomes and conditional teacher targets on student-generated prefixes provide different information. This distinction motivates graded retention of failed cases rather than binary rejection. Our method derives its weighting signal specifically from successful-only learning and aggregates it into one weight per trajectory, preserving the original token-level supervision structure within each trajectory.

% =============================================================================
\section{Preliminaries}
% =============================================================================
\label{sec:preliminary}
\subsection{On-Policy Distillation}

On-policy distillation (OPD) trains a student policy using states induced by the student's own generations while querying a stronger teacher for dense supervision on those states. Let $x \sim \mathcal{D}$ denote a problem sampled from the training distribution and let the student policy $\pi_\theta$ generate a response
\begin{equation}
\tau=(a_1,\ldots,a_T)\sim \pi_\theta(\cdot\mid x).
\end{equation}
At step $t$, the corresponding state is
\begin{equation}
s_t=(x,a_{<t}),
\end{equation}
which contains the original problem and the student-generated prefix.

A teacher policy $\pi_T$ is evaluated on the student-visited states
$s_t$. Standard OPD trains the student by minimizing the reverse
Kullback--Leibler (KL) divergence from the student policy to the teacher
distribution at these states:
\begin{equation}
\mathcal{L}_{\mathrm{OPD}}(\theta)
=
\mathbb{E}_{x\sim\mathcal{D},\,
\tau\sim\pi_\theta(\cdot\mid x)}
\left[
\sum_{t=1}^{T}
D_{\mathrm{KL}}
\left(
\pi_\theta(\cdot\mid s_t)
\,\|\, 
\pi_T(\cdot\mid s_t)
\right)
\right].
\end{equation}
The reverse-KL objective encourages the student to match the teacher
distribution on states induced by the student's generations.
Practical implementations may use sampled-token estimators of this
objective, but the underlying distillation target remains the same.
The defining feature of OPD is therefore that the distillation states
are induced by the student itself, unlike off-policy distillation.

\subsection{Online and Offline On-Policy Distillation}

Online and offline OPD differ in when student trajectories and teacher supervision are collected.
In online OPD, trajectory collection and optimization are interleaved. At training iteration $k$, the current student policy $\pi_{\theta_k}$ generates new trajectories,
\begin{equation}
\tau^{(k)}\sim\pi_{\theta_k}(\cdot\mid x),
\end{equation}
and the teacher is queried on the corresponding student-visited states. The resulting supervision updates the student before the next generation round, so both the student state distribution and teacher queries evolve throughout training. This tight coupling preserves on-policy coverage but incurs substantial computational cost from repeated student generation and teacher inference.

In offline OPD, trajectory collection is separated from optimization.
A fixed student checkpoint $\rho$ first generates a collection of trajectories,
\begin{equation}
\tau_i \sim \rho(\cdot \mid x_i),
\end{equation}
and the teacher distributions on the resulting states are computed and stored
once. Training then repeatedly optimizes the reverse-KL objective on this fixed
collection,
\begin{equation}
\mathcal{L}_{\mathrm{offline}}(\theta)
=
\sum_{i=1}^{N}
\sum_{t=1}^{T_i}
D_{\mathrm{KL}}
\left(
\pi_\theta(\cdot \mid s_{i,t})
\,\|\, 
\pi_T(\cdot \mid s_{i,t})
\right).
\end{equation}
By removing repeated generation and online teacher inference,
offline OPD substantially reduces training cost \citep{wu2026lightning}.
This is particularly attractive when GPU resources are limited or the teacher
itself requires substantial inference resources. However, reusing the same
teacher supervision also makes its quality more consequential, which motivates
our study of imperfect teachers.

% =============================================================================
\section{Methods}
\label{sec:theory}
\begin{figure}[t]
\centering
\includegraphics[width=\linewidth]{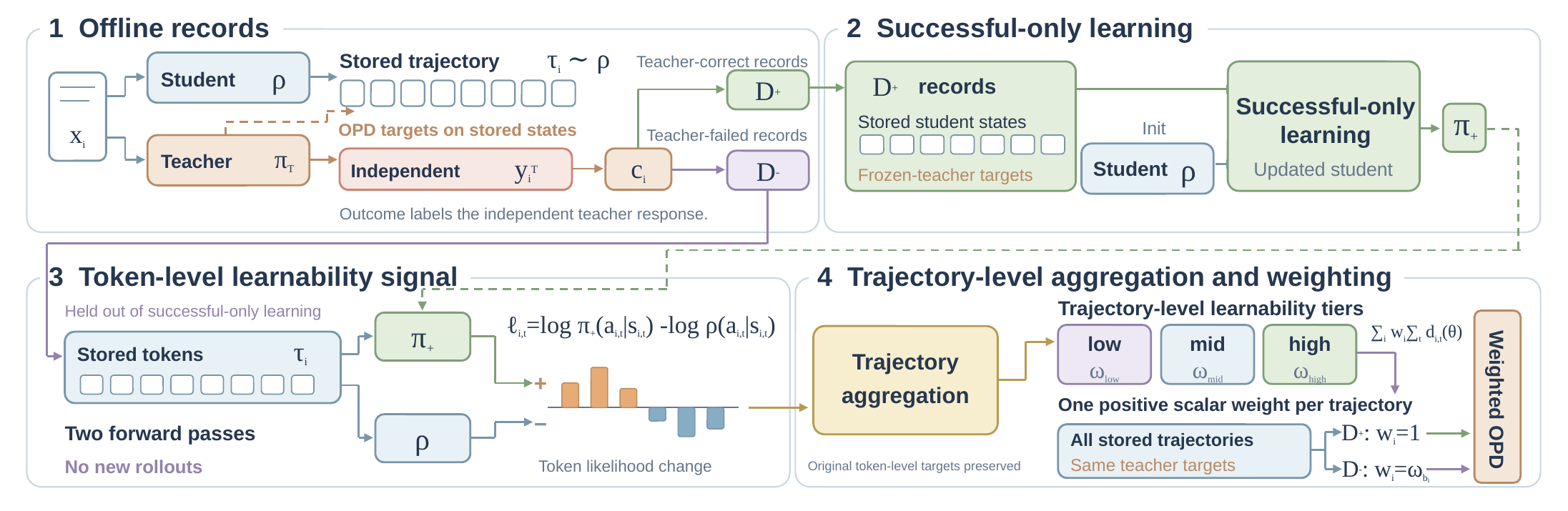}
\caption{Learnability-Weighted Distillation (LWD) for offline OPD.
(1) We collect student trajectories, frozen-teacher targets, and outcomes of independently generated teacher responses.
(2) Teacher-successful records train a reference checkpoint $\pi_+$ from $\rho$.
(3) Stored trajectories from teacher-failed problems are rescored under both checkpoints, and their token-level likelihood changes are summarized by the trajectory mean and variation.
(4) The resulting tiers assign bounded positive trajectory weights for final offline distillation. Scoring uses two forward passes and no new rollouts. The token profile is schematic; the displayed weights are realized Code values.}
\label{fig:method}
\end{figure}

% =============================================================================
We now describe \emph{Learnability-Weighted Distillation (LWD)}, following
the four stages illustrated in Figure~\ref{fig:method}. We first construct records containing student trajectories, teacher supervision, and teacher outcomes. We then learn a reference update from
teacher-successful problems and use it to define a token-level
\emph{learnability signal} on trajectories from teacher-failed problems. Finally, we
aggregate this signal into trajectory-level learnability tiers and use the
weights to scale the original OPD objective.

% -----------------------------------------------------------------------------
\subsection{Offline Records and Teacher Outcomes}
\label{sec:method-setup}
% -----------------------------------------------------------------------------

Let $\mathcal{D}=\{x_i\}_{i=1}^{N}$ denote the training problem set, and
let $\rho$ denote the fixed student policy used to collect the offline
trajectories. For each problem $x_i$, we store
\begin{equation}
\tau_i
=
(a_{i,1},\ldots,a_{i,T_i})
\sim
\rho(\cdot\mid x_i),
\end{equation}
where the state at token $t$ is
\begin{equation}
s_{i,t}
=
(x_i,a_{i,<t}).
\end{equation}

A frozen teacher $\pi_T$ provides the token-level supervision used by the
offline OPD objective. Following the reverse-KL formulation in
Section~\ref{sec:preliminary}, we denote the token-level loss used on
the stored state--action pair $(s_{i,t},a_{i,t})$ by
$d_{i,t}(\theta)$.

Separately, the teacher generates one response $y_i^T$ to each problem.
A domain-specific success predicate defines
\begin{equation}
c_i
=
\mathbf{1}
\left[
\Succ(y_i^T,x_i)
\right],
\end{equation}
which partitions the records into
\begin{equation}
\mathcal{D}^{+}
=
\{i:c_i=1\},
\qquad
\mathcal{D}^{-}
=
\{i:c_i=0\}.
\end{equation}

Importantly, $c_i$ records the outcome of an independently generated
teacher response. It labels neither the correctness of the stored student trajectory $\tau_i$ nor the quality of every teacher target
$\pi_T(\cdot\mid s_{i,t})$ on the student-visited states. Thus, $c_i$ is a problem-level outcome signal rather than a token-level judgment of the OPD supervision.

% -----------------------------------------------------------------------------
\subsection{Successful-Only Reference}
\label{sec:successful-reference}
% -----------------------------------------------------------------------------

We next train the student only on records whose teacher response succeeds.
Starting from the trajectory-collecting policy
$\rho=\pi_{\theta_\rho}$, we apply the same distillation procedure using
only $\mathcal{D}^{+}$:
\begin{equation}
\theta_{+}
=
\operatorname{Train}
\left(
\theta_\rho,\,
\{\tau_i,\pi_T\}_{i\in\mathcal{D}^{+}}
\right),
\qquad
\pi_{+}
=
\pi_{\theta_{+}}.
\label{eq:successful-checkpoint}
\end{equation}

We refer to the update from $\rho$ to $\pi_{+}$ as
\emph{successful-only learning}. The trajectories remain student-generated;
only the subset used for this auxiliary distillation is selected by the
teacher outcome. Thus, $\pi_{+}$ captures the model change induced by
learning from teacher supervision on problems accompanied by observed
teacher success.
Records in $\mathcal{D}^{-}$ are excluded from this update. They can
therefore be used to ask how successful-only learning changes behavior
on teacher-failed problems.

% -----------------------------------------------------------------------------
\subsection{Token-Level Learnability Signal}
\label{sec:learnability-signal}
% -----------------------------------------------------------------------------

For each trajectory $\tau_i$ with $i\in\mathcal{D}^{-}$, we score the
same stored tokens under both $\rho$ and $\pi_{+}$. We define
\begin{equation}
\ell_{i,t}
=
\log \pi_{+}(a_{i,t}\mid s_{i,t})
-
\log \rho(a_{i,t}\mid s_{i,t}).
\label{eq:token-logratio}
\end{equation}

We use $\ell_{i,t}$ as an operational token-level
\emph{learnability signal}. A positive value means that successful-only
learning increases the likelihood of the observed student token, whereas
a negative value means that the same update decreases it. The signal does
not label the token as correct and does not estimate the utility of the
teacher target at that state; it measures how the observed student
behavior responds to successful-only learning.
For interpretation, averaging the unsmoothed token-level signal gives
\begin{equation}
\bar{\ell}_i
=
\frac{1}{T_i}
\sum_{t=1}^{T_i}
\ell_{i,t}
=
\frac{1}{T_i}
\log
\frac{
\pi_{+}(\tau_i\mid x_i)
}{
\rho(\tau_i\mid x_i)
}.
\label{eq:sequence-logratio}
\end{equation}
The second equality is exact because both policies score the same
autoregressive sequence.

\paragraph{First-order interpretation.}
Let
\begin{equation}
f_i(\theta)
=
\frac{1}{T_i}
\log \pi_\theta(\tau_i\mid x_i),
\qquad
\Delta\theta
=
\theta_{+}-\theta_\rho.
\end{equation}
A first-order expansion around $\theta_\rho$ gives
\begin{equation}
\bar{\ell}_i
=
\Delta\theta^\top
\nabla_\theta f_i(\theta_\rho)
+
O\!\left(\|\Delta\theta\|^2\right).
\label{eq:local-alignment}
\end{equation}
Thus, to first order, the average signal reflects alignment between the
successful-only update and the direction that increases the likelihood of
the stored trajectory. This interpretation is approximate, whereas
Eq.~\eqref{eq:sequence-logratio} is exact.

% -----------------------------------------------------------------------------
\subsection{Trajectory-Level Learnability Weighting}
\label{sec:score-to-weight}
% -----------------------------------------------------------------------------

The token-level learnability signal should determine how much OPD supervision each record receives. In principle, it can be applied at the token, span, or trajectory level. However, the learnability signal and the OPD loss describe different quantities. The former measures how successful-only learning changes the likelihood of the student token actually generated, whereas the latter trains the student toward the teacher distribution at the same state. A low learnability signal may therefore occur precisely where the student deviates and corrective teacher supervision is still useful.

For this reason, directly reweighting individual tokens or spans can unnecessarily reshape the original token-level OPD supervision. We instead aggregate the local signal into a trajectory-level weight, which adjusts the overall supervision strength while preserving the relative token-level structure within the trajectory. This trajectory-level design also performs best in our ablations.

We first smooth the token-level signal along each response to reduce local
fluctuations. Let $\widetilde{\ell}_{i,t}$ denote the smoothed signal.
We summarize each trajectory by its overall level
\begin{equation}
m_i
=
\frac{1}{T_i}
\sum_{t=1}^{T_i}
\widetilde{\ell}_{i,t},
\end{equation}
and its variation
\begin{equation}
v_i
=
\sqrt{
\frac{1}{T_i}
\sum_{t=1}^{T_i}
\left(
\widetilde{\ell}_{i,t}-m_i
\right)^2
}.
\end{equation}
The first statistic captures how strongly the trajectory is supported
overall by successful-only learning, while the second captures how stable
that support is across the response.

We standardize $m_i$ and $v_i$ within the teacher-failed subset,
\begin{equation}
z_i^{m}
=
\frac{m_i-\mu_m}{\sigma_m},
\qquad
z_i^{v}
=
\frac{v_i-\mu_v}{\sigma_v},
\end{equation}
and use their relative values to assign a broad ordinal tier
\begin{equation}
b_i
=
B\!\left(
z_i^{m},
z_i^{v}
\right)
\in
\{
\mathrm{low},
\mathrm{mid},
\mathrm{high}
\}.
\label{eq:tier-map}
\end{equation}
The three tiers correspond to relatively weak or unstable, intermediate,
and strong stable learnability, respectively. Exact smoothing choices,
standardization details, and tier boundaries are provided in the
implementation details.
Each tier is associated with a bounded positive trajectory weight,
\begin{equation}
0
<
\omega_{\mathrm{low}}
<
\omega_{\mathrm{mid}}
<
\omega_{\mathrm{high}}.
\end{equation}
Teacher-successful records retain their original weight, giving
\begin{equation}
w_i
=
\begin{cases}
1, & c_i=1,\\
\omega_{b_i}, & c_i=0.
\end{cases}
\label{eq:final-weight}
\end{equation}

The final learnability-weighted OPD objective is
\begin{equation}
\mathcal{L}_{\mathrm{LWD}}(\theta)
=
\sum_{i=1}^{N}
w_i
\sum_{t=1}^{T_i}
d_{i,t}(\theta).
\label{eq:lwd-objective}
\end{equation}
Because $w_i$ is shared across all tokens in $\tau_i$, LWD reallocates
supervision across trajectories while preserving the token-level teacher
targets and their relative structure within each trajectory.

% =============================================================================
\section{Experiments}
\label{sec:exp}
% =============================================================================

\begin{table}[t]
\centering
\small
\caption{Main results for distillation from Qwen3-8B to Qwen3 students at two scales.
We report avg@32 on mathematical reasoning benchmarks and avg@4 on
LiveCodeBench. Best results within each student scale are shown in bold.}
\label{tab:main}
\begin{tabular}{@{}llrrrrr@{}}
\toprule
Method & On/Off & AIME24 & AIME25 & HMMT25 & LCB v5 & LCB v6 \\
\midrule
\multicolumn{7}{@{}l}{\emph{Student: Qwen3-1.7B-Base}}\\
SFT  & --- & 6.5\% & 10.4\% & 3.7\% & 4.3\% & 9.0\% \\
OPD \citep{agarwal2024policy} & Online & 17.5\% & 20.5\% & 14.9\% & 7.5\% & \textbf{13.3\%} \\
RA-OPD \citep{gan2026teacher} & Online & 14.4\% & 18.9\% & 12.1\% & --- & --- \\
ReNIO \citep{lin2026renio} & Online & 19.3\% & 20.2\% & 13.0\% & \textbf{7.6\%} & 13.1\% \\
FiRe-OPD \citep{li2026filter} & Online & 16.4\% & 20.1\% & 12.9\% & 7.2\% & \textbf{13.3\%} \\
Lightning-OPD \citep{wu2026lightning} & Offline & 19.8\% & 24.8\% & 15.8\% & 5.8\% & 10.1\% \\
\textbf{LWD (ours)} & Offline & \textbf{21.4\%} & \textbf{25.6\%} & \textbf{16.8\%} & 6.3\% & 11.3\% \\
\midrule
\multicolumn{7}{@{}l}{\emph{Student: Qwen3-4B-Base}}\\
SFT & --- & 57.1\% & 52.1\% & 34.0\% & 34.7\% & 36.4\% \\
OPD \citep{agarwal2024policy} & Online & 65.4\% & 57.9\% & 39.9\% & 44.2\% & 39.3\% \\
RA-OPD \citep{gan2026teacher} & Online & 66.2\% & 53.5\% & 38.5\% & --- & --- \\
ReNIO \citep{lin2026renio}  & Online & 61.5\% & 53.2\% & 38.2\% & 38.0\% & 39.0\% \\
ExOPD \citep{yang2026learning} & Online & 61.0\% & 56.0\% & 34.4\% & 29.0\% & --- \\
FiRe-OPD \citep{li2026filter} & Online & 63.6\% & 55.6\% & 35.8\% & 41.0\% & 39.6\% \\
%RA-OPD & Online & \TBD & \TBD & \TBD & \TBD & \TBD \\
%Lightning-OPD (reported) & Offline & 68.1\% & 58.4\% & 39.8\% & 42.8\% & 40.3\% \\
Lightning-OPD \citep{wu2026lightning}  & Offline & 66.4\% & 58.0\% & 40.5\% & 42.7\% & 40.4\% \\
\textbf{LWD (ours)} & Offline & \textbf{69.1\%} & \textbf{60.0\%} & \textbf{42.1\%} & \textbf{44.5\%} & \textbf{42.1\%} \\
\bottomrule
\end{tabular}
\end{table}

\noindent \textbf{Datasets.}
We evaluate our method on mathematical reasoning and code generation.
For mathematical reasoning, we train on DAPO-Math-17K
\citep{yu2026dapo} and evaluate on AIME 2024, AIME 2025, and HMMT 2025.
For code generation, we train on the 30K function-generation subset of
EpiCoder-300K used by Lightning-OPD
\citep{DBLP:conf/icml/WangL000HGHX0S025,wu2026lightning} and evaluate on
LiveCodeBench (LCB) v5 and v6
\citep{DBLP:conf/iclr/JainHGLYZWSSS25}.

\noindent \textbf{Training and implementation.}
We follow the Lightning-OPD protocol \citep{wu2026lightning} where applicable and evaluate distillation from Qwen3-8B to Qwen3-4B-Base and
Qwen3-1.7B-Base. We chose Qwen because it was widely used and among the best-performing open-source models. For fair comparison, all methods within each student scale use the same initialization, training data, GPU platform, optimization budget, and number of training steps. We report only the fixed final checkpoint, with no checkpoint selection or cherry-picking. More details are provided in Appendix~\ref{supp:implementation}.

\noindent \textbf{Evaluation protocol.}
Following Lightning-OPD \citep{wu2026lightning}, we report avg@32 for
AIME 2024, AIME 2025, and HMMT 2025, and avg@4 for LiveCodeBench v5 and v6,
where avg@$K$ is the fraction of successful generations over $K$ samples.
All methods use the same GPU platform, vLLM and Hugging Face
Transformers versions, seed, and tensor-parallel size. We evaluate with
temperature $0.6$, top-$p$ $0.95$, and maximum lengths of 32K tokens for
mathematics and 40K for code.

\noindent \textbf{Baselines.}
We compare against standard OPD \citep{agarwal2024policy}, selective and
reweighting-based OPD methods including ReNIO \citep{lin2026renio} and
FiRe-OPD \citep{li2026filter}, reliability-aware RA-OPD
\citep{gan2026teacher}, and ExOPD \citep{yang2026learning}. We use
Lightning-OPD \citep{wu2026lightning} as the primary offline baseline,
since our method operates in the same offline setting and is designed to
improve how its reused teacher supervision is weighted.

\subsection{Main results.}
Table~\ref{tab:main} compares our method with both online and offline OPD
baselines across mathematical reasoning and code generation.
At the 4B scale, LWD consistently improves the protocol-matched
Lightning-OPD baseline on all five benchmarks, with gains of $+2.7$ points
on AIME24, $+2.0$ on AIME25, $+1.6$ on HMMT25, $+1.8$ on LCB v5, and
$+1.7$ on LCB v6. LWD also consistently outperforms the reliability-aware
RA-OPD baseline across all three mathematics benchmarks. It achieves the
best result among all compared methods on every benchmark. Notably, its
AIME24 result of $69.1\%$ also exceeds the $68.1\%$ originally reported by
Lightning-OPD, despite all controlled comparisons in our main table using
our own protocol-matched reproduction.

The same trend largely transfers to the smaller 1.7B student.
LWD improves the offline Lightning-OPD baseline on all five benchmarks
and achieves the strongest performance on all three mathematics
benchmarks. On code generation, online OPD variants remain slightly
stronger at this smaller model scale, but LWD substantially narrows the
gap while retaining the lower-cost offline training.
Overall, the gains are consistent across domains and student scales.
The strong 4B results in particular show that learnability-guided
retention of teacher-failed supervision can match or outperform online
OPD variants without requiring online teacher interaction during training.

\noindent \textbf{Training Efficiency.}
Despite the additional successful-only distillation stage, LWD preserves the
computational advantage of offline OPD over online alternatives. It requires
$2$ GPUs, $10.98$ wall-clock hours, and $21.96$ GPU hours, compared with
$3$ GPUs and $12.09$--$15.90$ wall-clock hours and $36.28$--$47.69$ GPU hours
for the online selective OPD methods in our comparison.
The underlying training loop itself remains essentially unchanged: LWD and
Lightning-OPD require $2.14$ and $2.13$ minutes per optimization step,
respectively. The additional cost therefore comes primarily from the one-time
successful-only distillation stage rather than a slower optimization loop.
Compared with Lightning-OPD, this stage increases the total offline cost from
$10.64$ to $21.96$ GPU hours, while still retaining the lower two-GPU resource
requirement of offline training.
\begin{table}[t]
\centering
\small
\caption{
Training cost of offline and online OPD methods under the same hardware setting. LWD retains the two-GPU offline training setup while remaining below all online baselines in both wall-clock and GPU-hour cost. $n$A$m$T denotes $n$ actor GPUs and $m$ teacher GPUs.
}
\label{tab:training_efficiency}
\setlength{\tabcolsep}{5pt}
\renewcommand{\arraystretch}{1.1}
\begin{tabular}{lcccc}
\toprule
Method & GPUs & Wall hours & GPU hours & Min./step \\
\midrule
Naive OPD      & 2A1T & $\sim$15.3 & $\sim$46.0 & $\sim$6.1 \\
ReNIO          & 2A1T & 15.90 & 47.69 & 6.15 \\
FiRe-OPD       & 2A1T & 12.09 & 36.28 & 4.84 \\
Lightning-OPD  & 2A   & 5.32  & 10.64 & 2.13 \\
\textbf{LWD (ours)}
               & 2A   & 10.98 & 21.96 & 2.14 \\
\bottomrule
\end{tabular}
\end{table}

\begin{table}[t]
\centering
\small
\caption{
Ablation on teacher-failed supervision. Reduced weighting outperforms both
full retention and removal, while trajectory-specific learnability weighting
provides strongest overall performance.
}
\label{tab:ablation_failed}
\begin{tabular}{@{}lrrrr@{}}
\toprule
Failed-case weight & AIME24 & AIME25 & HMMT25 & Avg. \\
\midrule
Full
    & 66.4\% & 58.0\% & 40.5\% & 55.0\% \\
Zero
    & 67.1\% & 59.6\% & 39.7\% & 55.5\% \\
Shared reduced
    & 68.7\% & 59.3\% & 41.1\% & 56.4\% \\
\textbf{Learnability-based}
    & \textbf{69.1}\% & \textbf{60.0}\% & \textbf{42.1}\% & \textbf{57.1}\% \\
\bottomrule
\end{tabular}
\end{table}

\subsection{Ablation Studies}

% \noindent \textbf{How should teacher-failed supervision be weighted?} Table~\ref{tab:ablation_failed} shows a progression in how teacher-failed supervision should be treated. Assigning full weight gives the lowest average performance, while setting the weight to zero provides only a modest improvement. This suggests that the teacher outcome is informative about supervision quality, but a failed teacher response is not a certificate that all OPD supervision should be discarded. Assigning a shared reduced weight improves the average from $55.5$ to $56.4$, showing that teacher-failed problems still contain useful distillation signal. Learnability-based weighting further improves all three benchmarks, reaching an average of $57.1$. Compared with the shared reduced weight, it gains $0.4$ points on AIME24, $0.7$ on AIME25, and $1.0$ on HMMT25. These results support our central hypothesis: teacher failure provides a useful coarse signal, but the remaining supervision value still varies across trajectories.
\noindent \textbf{How should teacher-failed supervision be weighted?} 
Table~\ref{tab:ablation_failed} shows that teacher-failed supervision should
neither be fully retained nor discarded. Setting its weight to zero slightly
improves over full weighting, suggesting that teacher failure is informative
but does not make the associated OPD supervision useless. A shared reduced
weight further improves the average from $55.5$ to $56.4$, while
learnability-based weighting reaches $57.1$ and improves all three benchmarks.
These results support our central hypothesis: teacher failure is a useful
coarse signal rather than a hard filter, but the remaining supervision value
still varies across trajectories within failed cases.

\noindent \textbf{At what granularity should learnability modify OPD?}
Table~\ref{tab:ablation_granularity} compares token-, span-, and
trajectory-level weighting. Fine-grained weighting can be effective on
individual benchmarks: dense span weighting reaches the best HMMT25 result
of $43.1$. However, no token- or span-level variant performs consistently
best across tasks. Continuous trajectory weighting gives the strongest
average among the continuous variants at $56.7$, while tiered trajectory
weighting further improves the average to $57.1$. These results support
using local learnability as evidence for adjusting the overall supervision
assigned to a trajectory rather than directly reshaping the token-level
OPD structure.

\begin{figure}[t]
\centering
\includegraphics[width=\linewidth]{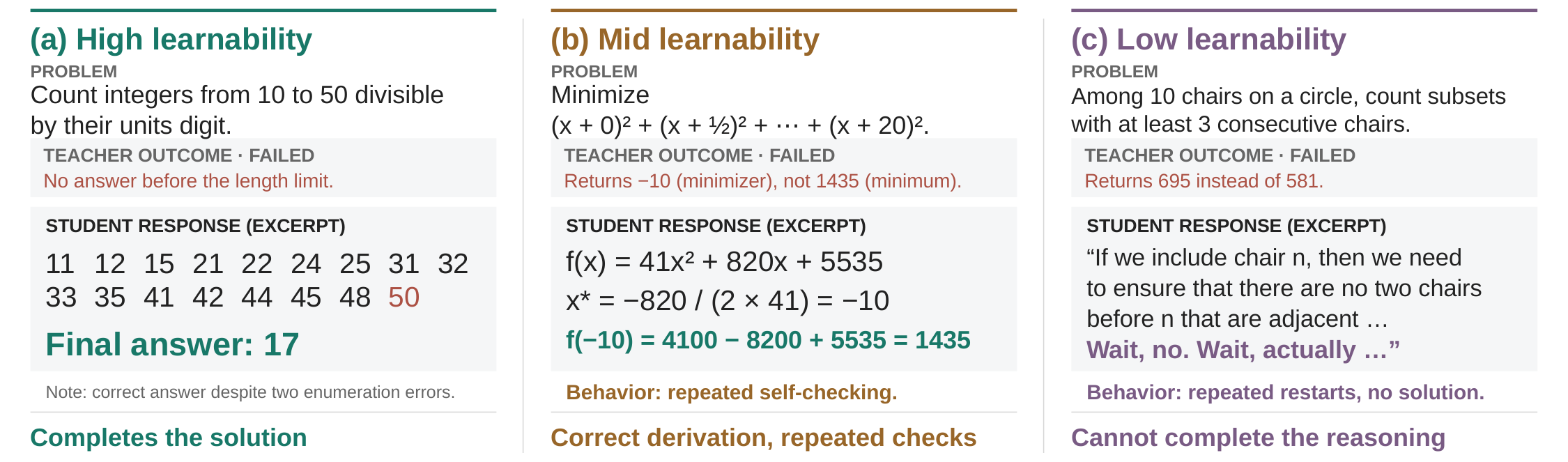}
\caption{Qualitative case study on mathematical reasoning under teacher failure.
Although all examples are associated with failed independent teacher responses,
their student trajectories exhibit substantially different learnability:
high-tier examples contain complete solutions, mid-tier examples mix correct
progress with repeated checking, and low-tier examples remain unresolved.}
\label{fig:learnability-case-study}
\end{figure}

\noindent \textbf{What do different learnability tiers capture?}
Figure~\ref{fig:learnability-case-study} qualitatively compares teacher-failed problems across learnability tiers. High-learnability trajectories can contain complete solutions despite teacher failure. Mid-learnability trajectories make correct progress but include repeated checking or unproductive deliberation, while low-learnability trajectories often stall or repeatedly restart. These examples show that the same teacher-failure label can correspond to different reasoning behaviors, supporting trajectory-specific supervision weights.

\begin{table}[t]
\centering
\small
\caption{
Ablation on learnability weighting granularity. Trajectory-level weighting
is more consistent than finer-grained alternatives, and tiered trajectory
weighting achieves the best performance.
}
\label{tab:ablation_granularity}
\begin{tabular}{@{}llrrrr@{}}
\toprule
Granularity & Weight form & AIME24 & AIME25 & HMMT25 & Avg. \\
\midrule
Token      & Sparse & 68.6\% & 58.0\% & 39.7\% & 55.4\% \\
Token    & Dense & 68.6\% & 60.5\% & 39.6\% & 56.2\% \\
Span      & Sparse     & 67.5\% & 59.5\% & 41.2\% & 56.1\% \\
Span      & Dense & 66.4\% & 60.4\% & \textbf{43.1}\% & 56.6\% \\
Trajectory  & Continuous & 67.8\% & 60.9\% & 41.5\% & 56.7\% \\
\textbf{Trajectory (Ours)} & \textbf{Tiered}
           & \textbf{69.1}\% & 60.0\% & 42.1\% & \textbf{57.1}\% \\
\bottomrule
\end{tabular}
\end{table}

% =============================================================================
\section{Conclusion}
% =============================================================================
We study what remains learnable from imperfect teacher supervision in
offline on-policy distillation, where a failed teacher rollout does not imply
that every conditional target on student-visited states is useless.
Directly estimating recoverability through repeated continuations would
largely sacrifice offline efficiency, so we use successful-only learning to
construct a cheap learnability signal and aggregate it into trajectory-level
weights for the original OPD objective.
Across mathematical reasoning and code generation, our method improves a
protocol-matched offline OPD baseline and matches or exceeds online OPD
variants on multiple benchmarks, while requiring substantially fewer GPU
hours and fewer concurrently used GPUs.
These results suggest that teacher failure is a useful coarse signal rather
than a binary decision about whether supervision should be retained, and that
successful-only learning can distinguish how much to retain for each trajectory. This allows LWD to retain supervision from teacher-failed problems without treating cases as equally reliable.
By retaining failed-case supervision, LWD turns teacher failure from a binary filter into a graded signal for allocating distillation strength across trajectories.

% % =============================================================================
% \section*{The Use of Large Language Models}
% % =============================================================================
% In this work, we used generative AI tools to assist in developing and
% refining the conceptual framework and hypotheses; formulating mathematical
% claims; providing feedback on experimental design, controls, and ablations;
% assisting with parts of the method implementation and data-processing code;
% supporting qualitative case analysis and interpretation of experimental
% results; identifying and organizing relevant literature; creating and editing
% scientific figures; and drafting and editing portions of the manuscript.
% %
% All reported experiments were conducted on our own infrastructure,
% and all numerical results were audited by the authors. We did not use
% generative AI tools to generate synthetic training or evaluation data.
% We take responsibility for the final content of this work, including all
% text, claims, code, results, and artifacts.

\section*{Data Availability and Usage}

All data used in this work are publicly available open-source datasets.
No internal, proprietary, or user data is involved. For training, we use
DAPO-Math-17K for mathematical reasoning \citep{yu2026dapo} and a 30K
function-generation subset of EpiCoder for code instruction following
\citep{DBLP:conf/icml/WangL000HGHX0S025}. For evaluation, we report results
on AIME 2024, AIME 2025, HMMT 2025, and LiveCodeBench v5 and v6
\citep{DBLP:conf/iclr/JainHGLYZWSSS25}. The AIME and HMMT problems originate
from competitions organized by the Mathematical Association of America
and the Harvard--MIT Mathematics Tournament, respectively, and the
LiveCodeBench problems are collected from LeetCode, AtCoder, and Codeforces.
We access these evaluation problems solely through the corresponding public
benchmark releases and use them exclusively for evaluation.
Figure~\ref{fig:learnability-case-study} reproduces selected mathematical
evaluation problems for qualitative analysis. Each dataset is used in
accordance with its original license, and we cite the original sources
accordingly.

\bibliography{iclr2027_conference}

@article{sang2026crisp,
  title={Crisp: Compressed reasoning via iterative self-policy distillation},
  author={Sang, Hejian and Xu, Yuanda and Zhou, Zhengze and He, Ran and Wang, Zhipeng and Sun, Jiachen},
  journal={arXiv preprint arXiv:2603.05433},
  year={2026}
}

@article{xu2026tip,
  title={Tip: Token importance in on-policy distillation},
  author={Xu, Yuanda and Sang, Hejian and Zhou, Zhengze and He, Ran and Wang, Zhipeng and Geramifard, Alborz},
  journal={arXiv preprint arXiv:2604.14084},
  year={2026}
}

@inproceedings{agarwal2024policy,
  title={On-policy distillation of language models: Learning from self-generated mistakes},
  author={Agarwal, Rishabh and Vieillard, Nino and Zhou, Yongchao and Stanczyk, Piotr and Ramos Garea, Sabela and Geist, Matthieu and Bachem, Olivier},
  booktitle={International Conference on Learning Representations},
  volume={2024},
  pages={21246--21263},
  year={2024}
}

@article{wu2026lightning,
  title={Lightning opd: Efficient post-training for large reasoning models with offline on-policy distillation},
  author={Wu, Yecheng and Han, Song and Cai, Hai},
  journal={arXiv preprint arXiv:2604.13010},
  year={2026}
}

@article{li2026rethinking,
  title={Rethinking on-policy distillation of large language models: Phenomenology, mechanism, and recipe},
  author={Li, Yaxuan and Zuo, Yuxin and He, Bingxiang and Zhang, Jinqian and Xiao, Chaojun and Qian, Cheng and Yu, Tianyu and Gao, Huan-ang and Yang, Wenkai and Liu, Zhiyuan and others},
  journal={arXiv preprint arXiv:2604.13016},
  year={2026}
}

@article{lin2026renio,
  title={ReNIO: Reweighting Negative Trajectory Importance for LLM On-Policy Distillation},
  author={Lin, Chen and Chen, Kedi and Zhang, Wei},
  journal={arXiv preprint arXiv:2606.23104},
  year={2026}
}

@article{li2026filter,
  title={Filter, then reweight: Rethinking optimization granularity in on-policy distillation},
  author={Li, Yuying and Zheng, Leqi and Yu, Yongzi and Zhou, Wenrui and Zhong, Xuchang and Hu, Xing and Jin, Jing and Yuan, Hangjie and Feng, Tao},
  journal={arXiv preprint arXiv:2606.02684},
  year={2026}
}

@article{gan2026teacher,
  title={When Teacher Guidance Misleads: Reward-Aligned On-Policy Distillation},
  author={Gan, Siyuan and Li, Yuhan and Wang, Xiran and Meng, Linjian and Wang, Boyan and Zhao, Zhen and Huo, Jing and Gao, Yang},
  journal={arXiv preprint arXiv:2608.27960},
  year={2026}
}

@article{zhu2026reorder,
  title={ReOrder-OPD: Reliability-Aware Prompt Ordering for On-Policy Distillation},
  author={Zhu, Ximo and Liu, Ruiqi and Wang, Rong and Wu, Ping and Zheng, Xiang and Xu, Wenzhuo and Yao, Xubin and Yan, Zhiyuan and Li, Bo and Gao, Jun and others},
  journal={arXiv preprint arXiv:2608.10905},
  year={2026}
}

@article{zhang2026verify,
  title={Verify Before You Distill: Prompt-Level Teacher Gating for On-Policy Distillation},
  author={Zhang, Zhiwei and Sun, Zechen and Zhao, Fei and Peng, Kang and Liang, Bin and Deng, Huayu and Hu, Yao and Wong, Kam-Fai and Chuan, Mu},
  journal={arXiv preprint arXiv:2609.02998},
  year={2026}
}

@article{yang2026learning,
  title={Learning beyond teacher: Generalized on-policy distillation with reward extrapolation},
  author={Yang, Wenkai and Liu, Weijie and Xie, Ruobing and Yang, Kai and Yang, Saiyong and Lin, Yankai},
  journal={arXiv preprint arXiv:2602.12125},
  year={2026}
}

@article{liu2026teacher,
  title={When Are Teacher Tokens Reliable? Position-Weighted On-Policy Self-Distillation for Reasoning},
  author={Liu, Xiaogeng and Wang, Xinyan and Ma, Yingzi and Zhang, Yechao and Xiao, Chaowei},
  journal={arXiv preprint arXiv:2605.21606},
  year={2026}
}

@article{hinton2015distilling,
  title={Distilling the knowledge in a neural network},
  author={Hinton, Geoffrey and Vinyals, Oriol and Dean, Jeff},
  journal={arXiv preprint arXiv:1503.02531},
  year={2015}
}

@inproceedings{kim2016sequence,
  title={Sequence-level knowledge distillation},
  author={Kim, Yoon and Rush, Alexander M},
  booktitle={Proceedings of the 2016 conference on empirical methods in natural language processing},
  pages={1317--1327},
  year={2016}
}

@inproceedings{furlanello2018born,
  title={Born again neural networks},
  author={Furlanello, Tommaso and Lipton, Zachary and Tschannen, Michael and Itti, Laurent and Anandkumar, Anima},
  booktitle={International conference on machine learning},
  pages={1607--1616},
  year={2018},
  organization={PMLR}
}

@inproceedings{gu2024minillm,
  title={Minillm: Knowledge distillation of large language models},
  author={Gu, Yuxian and Dong, Li and Wei, Furu and Huang, Minlie},
  booktitle={International Conference on Learning Representations},
  volume={2024},
  pages={32694--32717},
  year={2024}
}

@inproceedings{hsieh2023distilling,
  title={Distilling step-by-step! outperforming larger language models with less training data and smaller model sizes},
  author={Hsieh, Cheng-Yu and Li, Chun-Liang and Yeh, Chih-Kuan and Nakhost, Hootan and Fujii, Yasuhisa and Ratner, Alex and Krishna, Ranjay and Lee, Chen-Yu and Pfister, Tomas},
  booktitle={Findings of the association for computational linguistics: ACL 2023},
  pages={8003--8017},
  year={2023}
}

@inproceedings{shridhar2023distilling,
  title={Distilling reasoning capabilities into smaller language models},
  author={Shridhar, Kumar and Stolfo, Alessandro and Sachan, Mrinmaya},
  booktitle={Findings of the Association for Computational Linguistics: ACL 2023},
  pages={7059--7073},
  year={2023}
}

@inproceedings{li2023symbolic,
  title={Symbolic chain-of-thought distillation: Small models can also “think” step-by-step},
  author={Li, Liunian Harold and Hessel, Jack and Yu, Youngjae and Ren, Xiang and Chang, Kai-Wei and Choi, Yejin},
  booktitle={Proceedings of the 61st Annual Meeting of the Association for Computational Linguistics (Volume 1: Long Papers)},
  pages={2665--2679},
  year={2023}
}

@article{mukherjee2023orca,
  title={Orca: Progressive learning from complex explanation traces of gpt-4},
  author={Mukherjee, Subhabrata and Mitra, Arindam and Jawahar, Ganesh and Agarwal, Sahaj and Palangi, Hamid and Awadallah, Ahmed},
  journal={arXiv preprint arXiv:2306.02707},
  year={2023}
}

@article{feng2024keypoint,
  title={Keypoint-based progressive chain-of-thought distillation for llms},
  author={Feng, Kaituo and Li, Changsheng and Zhang, Xiaolu and Zhou, Jun and Yuan, Ye and Wang, Guoren},
  journal={arXiv preprint arXiv:2405.16064},
  year={2024}
}

@inproceedings{li2024mode,
  title={Mode-cotd: Chain-of-thought distillation for complex reasoning tasks with mixture of decoupled lora-experts},
  author={Li, Xiang and He, Shizhu and Wu, Jiayu and Yang, Zhao and Xu, Yao and jun Jun, Yang and Liu, Haifeng and Liu, Kang and Zhao, Jun},
  booktitle={Proceedings of the 2024 joint international conference on computational linguistics, language resources and evaluation (LREC-COLING 2024)},
  pages={11475--11485},
  year={2024}
}

@article{yang2025survey,
  title={Survey on knowledge distillation for large language models: methods, evaluation, and application},
  author={Yang, Chuanpeng and Zhu, Yao and Lu, Wang and Wang, Yidong and Chen, Qian and Gao, Chenlong and Yan, Bingjie and Chen, Yiqiang},
  journal={ACM Transactions on Intelligent Systems and Technology},
  volume={16},
  number={6},
  pages={1--27},
  year={2025},
  publisher={ACM New York, NY}
}

@article{zelikman2022star,
  title={Star: Bootstrapping reasoning with reasoning},
  author={Zelikman, Eric and Wu, Yuhuai and Mu, Jesse and Goodman, Noah},
  journal={Advances in Neural Information Processing Systems},
  volume={35},
  pages={15476--15488},
  year={2022}
}

@article{cobbe2021training,
  title={Training verifiers to solve math word problems},
  author={Cobbe, Karl and Kosaraju, Vineet and Bavarian, Mohammad and Chen, Mark and Jun, Heewoo and Kaiser, Lukasz and Plappert, Matthias and Tworek, Jerry and Hilton, Jacob and Nakano, Reiichiro and others},
  journal={arXiv preprint arXiv:2110.14168},
  year={2021}
}

@inproceedings{lightman2024let,
  title={Let's verify step by step},
  author={Lightman, Hunter and Kosaraju, Vineet and Burda, Yuri and Edwards, Harrison and Baker, Bowen and Lee, Teddy and Leike, Jan and Schulman, John and Sutskever, Ilya and Cobbe, Karl},
  booktitle={International Conference on Learning Representations},
  volume={2024},
  pages={39578--39601},
  year={2024}
}

@article{yuan2023scaling,
  title={Scaling relationship on learning mathematical reasoning with large language models},
  author={Yuan, Zheng and Yuan, Hongyi and Li, Chengpeng and Dong, Guanting and Lu, Keming and Tan, Chuanqi and Zhou, Chang and Zhou, Jingren},
  journal={arXiv preprint arXiv:2308.01825},
  year={2023}
}

@inproceedings{deng2026beyond,
  title={Beyond Rejection Sampling: Trajectory Fusion for Scaling Mathematical Reasoning},
  author={Deng, Jie and Tong, Hanshuang and Li, Jun and Liang, Shining and Wu, Ning and Li, Hongzhi and Xie, Yutao},
  booktitle={Findings of the Association for Computational Linguistics: ACL 2026},
  pages={7943--7959},
  year={2026}
}

@article{yu2026dapo,
  title={Dapo: An open-source llm reinforcement learning system at scale},
  author={Yu, Qiying and Zhang, Zheng and Zhu, Ruofei and Yuan, Yufeng and Zuo, Xiaochen and Yue, Yu and Dai, Weinan and Fan, Tiantian and Liu, Gaohong and Liu, Lingjun and others},
  journal={Advances in Neural Information Processing Systems},
  volume={38},
  pages={113222--113244},
  year={2026}
}

@inproceedings{DBLP:conf/icml/BengioLCW09,
  author       = {Yoshua Bengio and
                  J{\'{e}}r{\^{o}}me Louradour and
                  Ronan Collobert and
                  Jason Weston},
  editor       = {Andrea Pohoreckyj Danyluk and
                  L{\'{e}}on Bottou and
                  Michael L. Littman},
  title        = {Curriculum learning},
  booktitle    = {Proceedings of the 26th Annual International Conference on Machine
                  Learning, {ICML} 2009, Montreal, Quebec, Canada, June 14-18, 2009},
  series       = {{ACM} International Conference Proceeding Series},
  volume       = {382},
  pages        = {41--48},
  publisher    = {{ACM}},
  year         = {2009},
  url          = {https://doi.org/10.1145/1553374.1553380},
  doi          = {10.1145/1553374.1553380},
  bibsource    = {dblp computer science bibliography, https://dblp.org}
}

@inproceedings{DBLP:conf/nips/KumarPK10,
  author       = {M. Pawan Kumar and
                  Benjamin Packer and
                  Daphne Koller},
  editor       = {John D. Lafferty and
                  Christopher K. I. Williams and
                  John Shawe{-}Taylor and
                  Richard S. Zemel and
                  Aron Culotta},
  title        = {Self-Paced Learning for Latent Variable Models},
  booktitle    = {Advances in Neural Information Processing Systems 23: 24th Annual
                  Conference on Neural Information Processing Systems 2010. Proceedings
                  of a meeting held 6-9 December 2010, Vancouver, British Columbia,
                  Canada},
  pages        = {1189--1197},
  publisher    = {Curran Associates, Inc.},
  year         = {2010},
  url          = {https://proceedings.neurips.cc/paper/2010/hash/e57c6b956a6521b28495f2886ca0977a-Abstract.html},
  bibsource    = {dblp computer science bibliography, https://dblp.org}
}

@inproceedings{DBLP:conf/icml/JiangZLLF18,
  author       = {Lu Jiang and
                  Zhengyuan Zhou and
                  Thomas Leung and
                  Li{-}Jia Li and
                  Li Fei{-}Fei},
  editor       = {Jennifer G. Dy and
                  Andreas Krause},
  title        = {MentorNet: Learning Data-Driven Curriculum for Very Deep Neural Networks
                  on Corrupted Labels},
  booktitle    = {Proceedings of the 35th International Conference on Machine Learning,
                  {ICML} 2018, Stockholmsm{\"{a}}ssan, Stockholm, Sweden, July
                  10-15, 2018},
  series       = {Proceedings of Machine Learning Research},
  volume       = {80},
  pages        = {2309--2318},
  publisher    = {{PMLR}},
  year         = {2018},
  url          = {http://proceedings.mlr.press/v80/jiang18c.html},
  bibsource    = {dblp computer science bibliography, https://dblp.org}
}

@inproceedings{DBLP:conf/icml/RenZYU18,
  author       = {Mengye Ren and
                  Wenyuan Zeng and
                  Bin Yang and
                  Raquel Urtasun},
  editor       = {Jennifer G. Dy and
                  Andreas Krause},
  title        = {Learning to Reweight Examples for Robust Deep Learning},
  booktitle    = {Proceedings of the 35th International Conference on Machine Learning,
                  {ICML} 2018, Stockholmsm{\"{a}}ssan, Stockholm, Sweden, July
                  10-15, 2018},
  series       = {Proceedings of Machine Learning Research},
  volume       = {80},
  pages        = {4331--4340},
  publisher    = {{PMLR}},
  year         = {2018},
  url          = {http://proceedings.mlr.press/v80/ren18a.html},
  bibsource    = {dblp computer science bibliography, https://dblp.org}
}

@inproceedings{DBLP:conf/nips/HanYYNXHTS18,
  author       = {Bo Han and
                  Quanming Yao and
                  Xingrui Yu and
                  Gang Niu and
                  Miao Xu and
                  Weihua Hu and
                  Ivor W. Tsang and
                  Masashi Sugiyama},
  editor       = {Samy Bengio and
                  Hanna M. Wallach and
                  Hugo Larochelle and
                  Kristen Grauman and
                  Nicol{\`{o}} Cesa{-}Bianchi and
                  Roman Garnett},
  title        = {Co-teaching: Robust training of deep neural networks with extremely
                  noisy labels},
  booktitle    = {Advances in Neural Information Processing Systems 31: Annual Conference
                  on Neural Information Processing Systems 2018, NeurIPS 2018, December
                  3-8, 2018, Montr{\'{e}}al, Canada},
  pages        = {8536--8546},
  year         = {2018},
  url          = {https://proceedings.neurips.cc/paper/2018/hash/a19744e268754fb0148b017647355b7b-Abstract.html},
  bibsource    = {dblp computer science bibliography, https://dblp.org}
}

@article{DBLP:journals/corr/abs-1910-00177,
  author       = {Xue Bin Peng and
                  Aviral Kumar and
                  Grace Zhang and
                  Sergey Levine},
  title        = {Advantage-Weighted Regression: Simple and Scalable Off-Policy Reinforcement
                  Learning},
  journal      = {CoRR},
  volume       = {abs/1910.00177},
  year         = {2019},
  url          = {http://arxiv.org/abs/1910.00177},
  eprinttype   = {arXiv},
  eprint       = {1910.00177},
  bibsource    = {dblp computer science bibliography, https://dblp.org}
}

@inproceedings{DBLP:conf/nips/KumarZTL20,
  author       = {Aviral Kumar and
                  Aurick Zhou and
                  George Tucker and
                  Sergey Levine},
  editor       = {Hugo Larochelle and
                  Marc'Aurelio Ranzato and
                  Raia Hadsell and
                  Maria{-}Florina Balcan and
                  Hsuan{-}Tien Lin},
  title        = {Conservative Q-Learning for Offline Reinforcement Learning},
  booktitle    = {Advances in Neural Information Processing Systems 33: Annual Conference
                  on Neural Information Processing Systems 2020, NeurIPS 2020, December
                  6-12, 2020, virtual},
  year         = {2020},
  url          = {https://proceedings.neurips.cc/paper/2020/hash/0d2b2061826a5df3221116a5085a6052-Abstract.html},
  bibsource    = {dblp computer science bibliography, https://dblp.org}
}

@inproceedings{DBLP:conf/iclr/KostrikovNL22,
  author       = {Ilya Kostrikov and
                  Ashvin Nair and
                  Sergey Levine},
  title        = {Offline Reinforcement Learning with Implicit Q-Learning},
  booktitle    = {The Tenth International Conference on Learning Representations, {ICLR}
                  2022, Virtual Event, April 25-29, 2022},
  publisher    = {OpenReview.net},
  year         = {2022},
  url          = {https://openreview.net/forum?id=68n2s9ZJWF8},
  bibsource    = {dblp computer science bibliography, https://dblp.org}
}

@inproceedings{DBLP:conf/icml/WangL000HGHX0S025,
  author       = {Yaoxiang Wang and
                  Haoling Li and
                  Xin Zhang and
                  Jie Wu and
                  Xiao Liu and
                  Wenxiang Hu and
                  Zhongxin Guo and
                  Yangyu Huang and
                  Ying Xin and
                  Yujiu Yang and
                  Jinsong Su and
                  Qi Chen and
                  Scarlett Li},
  editor       = {Aarti Singh and
                  Maryam Fazel and
                  Daniel Hsu and
                  Simon Lacoste{-}Julien and
                  Felix Berkenkamp and
                  Tegan Maharaj and
                  Kiri Wagstaff and
                  Jerry Zhu},
  title        = {EpiCoder: Encompassing Diversity and Complexity in Code Generation},
  booktitle    = {Forty-second International Conference on Machine Learning, {ICML}
                  2025, Vancouver, BC, Canada, July 13-19, 2025},
  series       = {Proceedings of Machine Learning Research},
  volume       = {267},
  publisher    = {{PMLR} / OpenReview.net},
  year         = {2025},
  url          = {https://proceedings.mlr.press/v267/wang25bi.html},
  bibsource    = {dblp computer science bibliography, https://dblp.org}
}

@inproceedings{DBLP:conf/iclr/JainHGLYZWSSS25,
  author       = {Naman Jain and
                  King Han and
                  Alex Gu and
                  Wen{-}Ding Li and
                  Fanjia Yan and
                  Tianjun Zhang and
                  Sida Wang and
                  Armando Solar{-}Lezama and
                  Koushik Sen and
                  Ion Stoica},
  title        = {LiveCodeBench: Holistic and Contamination Free Evaluation of Large
                  Language Models for Code},
  booktitle    = {The Thirteenth International Conference on Learning Representations,
                  {ICLR} 2025, Singapore, April 24-28, 2025},
  publisher    = {OpenReview.net},
  year         = {2025},
  url          = {https://openreview.net/forum?id=chfJJYC3iL},
  bibsource    = {dblp computer science bibliography, https://dblp.org}
}
\bibliographystyle{iclr2027_conference}
\appendix

\section{Implementation Details and Evaluation Protocol}
\label{supp:implementation}

\subsection{Learnability Estimation and Weight Construction}

We use the notation of Section~\ref{sec:theory}. The trajectory-collecting
policy $\rho$ is also used as the initialization for both the auxiliary
successful-only model and the final student. To obtain $\pi_+$, we train
$\rho$ for 150 optimizer updates using only records in $\mathcal{D}^{+}$.

For each $i\in\mathcal{D}^{-}$, we score the stored response under $\rho$
and $\pi_+$ and compute the token-level learnability signal
\begin{equation}
\ell_{i,t}
=
\log \pi_{+}(a_{i,t}\mid s_{i,t})
-
\log \rho(a_{i,t}\mid s_{i,t}).
\end{equation}
We clip $\ell_{i,t}$ to $[-4,4]$ and smooth it with a 33-token moving
average. The resulting trajectory mean $m_i$ and variation $v_i$ are
standardized over $\mathcal{D}^{-}$ using response-length-weighted
statistics.

The learnability tiers are instantiated as
\begin{equation}
b_i =
\begin{cases}
\mathrm{low},  & z^m_i \leq -0.5 \ \lor\ z^v_i \geq 1.0,\\
\mathrm{high}, & z^m_i \geq 0.5 \ \land\ z^v_i < 1.0,\\
\mathrm{mid},  & \text{otherwise}.
\end{cases}
\end{equation}
The weights for the low, mid, and high tiers are fixed at $0.15$,
$0.25$, and $0.35$, respectively. Each trajectory's weight multiplies
all token-level OPD losses within that trajectory. Teacher-successful
records retain weight $1$.

\subsection{Final LWD Training}

The final LWD student is initialized from $\rho$ and trained on the full
offline collection. We use the same sampled-token OPD estimator as the
protocol-matched Lightning-OPD baseline; LWD only multiplies all token
losses from trajectory $\tau_i$ by its trajectory weight $w_i$, as in
Eq.~\eqref{eq:lwd-objective}.

\subsection{Training Configuration}

We instantiate the method with Qwen3-1.7B and Qwen3-4B SFT students and
a Qwen3-8B teacher. Mathematical reasoning uses DAPO-Math-17K
(17,398 prompts), while code training uses the EpiCoder function subset
(30,000 prompts; 29,943 uniquely aligned prompts in the 1.7B run).
For mathematical reasoning, $c_i$ is obtained by applying the DAPO answer
verifier to the independently generated Qwen3-8B teacher response. For code, where executable tests are unavailable
in the training records, teacher success is approximated by normal
termination and strict extraction of a complete Python program. These
code labels measure format validity rather than functional correctness.

Both the auxiliary policy and final student are trained for 150
optimizer updates. We use a maximum training response length of 4,096,
a global batch size of 256, and a constant learning rate of $2\times
10^{-6}$. Optimization uses Adam with $\beta_1=0.9$, $\beta_2=0.98$,
and weight decay $0.1$. Training uses tensor parallelism of size two,
sequence parallelism, full activation recomputation, dynamic batching,
and a maximum of 32,768 training tokens per GPU. Dropout is disabled.
Only the final checkpoint after 150 optimizer updates is converted to Hugging Face
format and used for evaluation.

\subsection{Mathematical Reasoning Evaluation}

We evaluate on AIME 2024, AIME 2025, and HMMT February 2025. Each
benchmark contains 30 problems. We generate 32 independent completions
per problem with temperature $0.6$, top-$p=0.95$, a maximum generation
length of 32,768 tokens, and reasoning enabled. The primary evaluation
uses seed 42. Additional decoding-seed runs change only the random seed.

The raw ChatML evaluation prompt is:
\begin{verbatim}
<|im_start|>user
Question: {problem}
Please reason step by step, and put your final answer within \boxed{}.
<|im_end|>
<|im_start|>assistant
\end{verbatim}

We first extract the last valid boxed expression and fall back to the
final ``Answer:'' line when necessary. We report avg@32 over all sampled
completions:
\begin{equation}
    \mathrm{avg@32}
    =
    \frac{1}{32N}
    \sum_{i=1}^{N}\sum_{j=1}^{32}
    \mathbb{I}\!\left[y_{i,j}\text{ is correct}\right].
\end{equation}

\subsection{Code Evaluation}

We evaluate code generation using the official LiveCodeBench execution
evaluator pinned to commit
\texttt{28fef95ea8c9f7a547c8329f2cd3d32b92c1fa24}.
Version 5 contains 167 problems and version 6 contains 175 problems.
We generate four completions per problem using temperature $0.6$,
top-$p=0.95$, seed 42, and a maximum generation length of 40,960 tokens.

We retain the official \texttt{CodeQwenInstruct} problem template and
append the following output contract:
\begin{verbatim}
Output protocol: You may reason inside <think>...</think>.
Immediately after </think>, output exactly one complete Python
program inside a single ```python ... ``` block. Do not restate
the problem, explain the solution, or emit any other text after
</think>. Your answer is invalid unless the complete program
appears in that code block.
\end{verbatim}

All completions are graded by executing the official benchmark tests.
We report avg@4 over the four independent generations:
\begin{equation}
    \mathrm{avg@4}
    =
    \frac{1}{4N}
    \sum_{i=1}^{N}\sum_{j=1}^{4}
    \mathbb{I}\!\left[y_{i,j}\text{ passes all tests}\right].
\end{equation}
We additionally record the fraction of completions from which a valid
Python program can be extracted as a diagnostic metric.

\section{Sensitivity to weighting choices.}
Table~\ref{tab:sensitivity} evaluates sensitivity to both the shared
failed-case weight and the learnability-tier boundaries. Varying the constant
weight from $0.15$ to $0.50$ yields similar average performance
($56.3$--$56.6$), although different settings favor different benchmarks.
This suggests that the benefit of retaining teacher-failed supervision is not
driven by a particular shared weight. At the same time, no constant weight
matches the average performance of learnability-based weighting, with the best
constant variant reaching $56.6$ compared with $57.1$ for LWD. This further
supports differentiating failed cases rather than assigning them a single
global weight.

We also vary the learnability-tier boundaries while keeping the middle-tier
weight fixed at $0.25$. Narrow and wide variants achieve averages of $56.4$
and $56.9$, respectively, while the default setting reaches $57.1$. The wide
variant remains close to the default despite changing which trajectories fall
into each tier, suggesting that the gain does not depend on one exact boundary
choice. Overall, these results indicate that LWD is reasonably robust to both
the global weighting strength and the tier construction, while trajectory-
specific weighting remains preferable to a shared failed-case weight.

\begin{table}[t]
\centering
\small
\caption{
Sensitivity to failed-case weighting and learnability-tier boundaries.
Constant weighting is robust across different failed-case weights, while
trajectory-specific tiered weighting performs best. For tier-boundary
variants, the middle-tier weight is fixed at $0.25$.
}
\label{tab:sensitivity}
\begin{tabular}{@{}lrrrr@{}}
\toprule
Variant & AIME24 & AIME25 & HMMT25 & Avg. \\
\midrule
\multicolumn{5}{@{}l}{\emph{Constant failed-case weight}} \\
Constant ($0.15$) & 67.5 & 60.0 & 42.4 & 56.6 \\
Constant ($0.25$) & 68.7 & 59.3 & 41.1 & 56.4 \\
Constant ($0.50$) & 66.8 & 
\textbf{61.1} & 41.1 & 56.3 \\
\midrule
\multicolumn{5}{@{}l}{\emph{Learnability-tier boundaries}} \\
Tiered (narrow) & 67.9 & 59.4 & 42.0 & 56.4 \\
Tiered (wide)   & 69.0 & 59.3 & \textbf{42.5} & 56.9 \\
\textbf{Tiered (ours)}
                & \textbf{69.1} & 60.0 & 42.1 & \textbf{57.1} \\
\bottomrule
\end{tabular}
\end{table}

\section{Details of Granularity Ablations in Table 4}
\label{supp:granularity}

We compare several ways of translating the token-level learnability signal
into OPD supervision. All variants use the same stored trajectories,
successful-only model, and token-level signal $\ell_{i,t}$; they differ in the granularity and form of the weighting applied to the original OPD loss.

\paragraph{Token-level weighting.}
Token-level variants apply learnability directly to individual OPD token
losses. The sparse variant uses the learnability signal to selectively
reweight a subset of token positions, while the dense variant assigns
learnability-dependent weights throughout the response. These variants make
the most local intervention, but can also alter the relative supervision
structure within a trajectory.

\paragraph{Span-level weighting.}
Span-level variants first aggregate neighboring token-level signals into
contiguous spans and assign a shared weight within each span. As at the token
level, the sparse variant selectively reweights only identified spans, whereas
the dense variant applies learnability-dependent weighting across the full
trajectory through span-level aggregation. Span weighting therefore reduces
token-level noise while still allowing supervision strength to vary within a
trajectory.

\paragraph{Trajectory-level weighting.}
For continuous trajectory weighting, the local learnability signal is
aggregated into one scalar weight for the entire trajectory, and this weight
multiplies all token-level OPD losses in that trajectory. Our final tiered
variant further replaces the continuous trajectory score with the low-, mid-,
and high-learnability tiers defined in Section~\ref{sec:score-to-weight}.
Both trajectory-level variants preserve the relative token-level OPD
supervision within each trajectory; they differ only in whether the resulting
trajectory weight is continuous or discretized into bounded tiers.

\section{Learnability-Tier Distribution}

Table~\ref{tab:lwd-tier-distribution} reports the distribution of
teacher-failed trajectories across the final low-, mid-, and
high-learnability tiers. The partition remains reasonably balanced across
both domains and student scales: each tier contains between $26.7\%$ and
$39.8\%$ of the teacher-failed trajectories. Thus, the tiering rule does not
collapse most failed examples into a single category, despite using the same
decision rule across Math and Code.

The saved trajectory statistics also follow the intended ordering. Across all
four settings, high-learnability trajectories have a less negative mean
learnability statistic $m_i$ than mid- and low-learnability trajectories,
while their within-trajectory variation $v_i$ is lower. For example, on the
4B Math setting, the mean $m_i$ changes from $-0.137$ for Low to $-0.107$
for Mid and $-0.073$ for High, while $v_i$ decreases from $0.070$ to
$0.062$ and $0.057$, respectively. Similar trends hold for both Code settings
and the 1.7B Math model. The median response length is $4096$ in all reported tiers, matching the training response limit. Every teacher-failed
record is assigned to exactly one tier, and the tier counts reconcile with
the final scoring artifacts.

\begin{table}[t]
\centering
\small
\setlength{\tabcolsep}{5pt}
\caption{
Distribution of teacher-failed trajectories across the final LWD
learnability tiers. Percentages are computed within teacher-failed records
for each domain and student scale.
}
\label{tab:lwd-tier-distribution}
\begin{tabular}{llrrrr}
\toprule
Domain & Student & Failed & Low $n$ (\%) & Mid $n$ (\%) & High $n$ (\%) \\
\midrule
Math & Qwen3-4B-Base   & 4,830 & 1,856 (38.43) & 1,388 (28.74) & 1,586 (32.84) \\
Math & Qwen3-1.7B-Base & 9,984 & 3,978 (39.84) & 2,984 (29.89) & 3,022 (30.27) \\
Code & Qwen3-4B-Base   & 4,758 & 1,677 (35.25) & 1,810 (38.04) & 1,271 (26.71) \\
Code & Qwen3-1.7B-Base & 4,758 & 1,642 (34.51) & 1,327 (27.89) & 1,789 (37.60) \\
\bottomrule
\end{tabular}
\end{table}

\section{Training-Data and Teacher-Outcome Statistics}

Table~\ref{tab:lwd-training-data} summarizes the final training records used
by LWD. Teacher outcome is obtained from an independently generated Qwen3-8B
response and does not label the correctness of the stored student trajectory.
For mathematics, the outcome is determined by answer verification, whereas
for code generation executable unit tests are unavailable in the training
records and we therefore use a strict response-validity proxy.

The final datasets contain between $15.9\%$ and $57.4\%$ teacher-failed
records, providing a substantial failed-case subset for learnability
estimation. A large fraction of stored student responses also reach the
4,096-token training limit, particularly for mathematical reasoning. Here,
``cap-hit'' means that the stored supervised response contains exactly
4,096 tokens; the training records do not provide a separate termination
reason, so this statistic should not be interpreted as an independently
verified truncation event.
All final training datasets reconcile with their corresponding training
runs, and no retained record has missing or mismatched token-level teacher
supervision. 

% \paragraph{Teacher-Outcome Statistics}
% \label{supp:data-statistics}

% Table~\ref{tab:lwd-training-data} summarizes the final training records used
% by LWD. Teacher outcome is computed from an independently generated
% Qwen3-8B response and is used only as a problem-level signal; it does not
% label the correctness of the stored student trajectory. Across the final
% training sets, teacher-failed records account for $15.9\%$--$57.4\%$ of the
% data, providing a substantial subset on which LWD differentiates the amount
% of retained OPD supervision.

\begin{table}[t]
\centering
\small
\caption{
Final training-data composition used by LWD. Percentages are computed over
the retained training records. Code teacher outcome is based on the strict
response-validity proxy described below.
}
\label{tab:lwd-training-data}
\begin{tabular}{llrrr}
\toprule
Domain & Student & Retained & Teacher success & Teacher failure \\
\midrule
Math & Qwen3-4B
& 17,398 & 12,568 (72.2\%) & 4,830 (27.8\%) \\
Math & Qwen3-1.7B
& 17,398 & 7,414 (42.6\%) & 9,984 (57.4\%) \\
Code & Qwen3-4B
& 30,000 & 25,242 (84.1\%) & 4,758 (15.9\%) \\
Code & Qwen3-1.7B
& 29,943 & 25,185 (84.1\%) & 4,758 (15.9\%) \\
\bottomrule
\end{tabular}
\end{table}

\paragraph{Math teacher-outcome labels.}
The 4B and 1.7B Math experiments use different saved versions of the
teacher-outcome verifier. The 4B run uses the later boxed-answer
re-verification, whereas the protocol-matched 1.7B run retains the historical
labels used during that experiment. We therefore treat the teacher-success
rates in Table~\ref{tab:lwd-training-data} as run-specific training statistics
and do not interpret their cross-scale difference as an effect of student
model size.

\paragraph{Code teacher-outcome proxy.}
Executable unit tests are unavailable for the EpiCoder training records.
We therefore define teacher success using a strict response-validity proxy:
the teacher must terminate normally and produce exactly one non-empty,
syntactically valid Python block after reasoning, without surrounding text.
This label measures response validity rather than functional correctness.
The 1.7B Code run contains 29,943 records because 57 examples with ambiguous
source alignment are removed during preprocessing.

\section{Scope and Limitations}

Our learnability signal is operational. It measures how stored student
behavior changes under successful-only learning rather than directly
estimating teacher-target correctness. We therefore use it as a relative
signal for adjusting trajectory-level supervision strength, while retaining
bounded positive weights for teacher-failed trajectories.

The method also depends on a coarse teacher-outcome signal. For mathematics,
this outcome can be verified against reference answers, whereas for code
training the available records do not provide executable unit tests and we
therefore use response validity as a proxy for teacher success. This proxy
does not establish functional correctness. Nevertheless, LWD only uses the
outcome to construct successful-only learning and never treats teacher failure
as a token-level correctness label.

Finally, the current evaluation focuses on mathematical reasoning and code
generation with Qwen3 students and a Qwen3-8B teacher. The exact tier
boundaries and weights are implementation choices rather than theoretically
unique values. Our sensitivity experiments show similar behavior across
multiple shared weights and alternative tier boundaries, but broader model
families, teacher strengths, and task domains remain useful directions for
future evaluation.

% % =============================================================================
% \section{Analysis and Limitations}
% \label{sec:analysis}
% % =============================================================================

\end{document}